\PassOptionsToPackage{unicode}{hyperref}
\PassOptionsToPackage{hyphens}{url}
\documentclass[
]{article}
\usepackage{amsmath,amssymb}
\usepackage{iftex}
\ifPDFTeX
  \usepackage[T1]{fontenc}
  \usepackage[utf8]{inputenc}
  \usepackage{textcomp} % provide euro and other symbols
\else % if luatex or xetex
  \usepackage{unicode-math} % this also loads fontspec
  \defaultfontfeatures{Scale=MatchLowercase}
  \defaultfontfeatures[\rmfamily]{Ligatures=TeX,Scale=1}
\fi
\usepackage{lmodern}
\ifPDFTeX\else
\fi
\IfFileExists{upquote.sty}{\usepackage{upquote}}{}
\IfFileExists{microtype.sty}{% use microtype if available
  \usepackage[]{microtype}
  \UseMicrotypeSet[protrusion]{basicmath} % disable protrusion for tt fonts
}{}
\makeatletter
\@ifundefined{KOMAClassName}{% if non-KOMA class
  \IfFileExists{parskip.sty}{%
    \usepackage{parskip}
  }{% else
    \setlength{\parindent}{0pt}
    \setlength{\parskip}{6pt plus 2pt minus 1pt}}
}{% if KOMA class
  \KOMAoptions{parskip=half}}
\makeatother
\usepackage{xcolor}
\usepackage{longtable,booktabs,array}
\usepackage{calc} % for calculating minipage widths
\usepackage{etoolbox}
\makeatletter
\patchcmd\longtable{\par}{\if@noskipsec\mbox{}\fi\par}{}{}
\makeatother
\IfFileExists{footnotehyper.sty}{\usepackage{footnotehyper}}{\usepackage{footnote}}
\makesavenoteenv{longtable}
\ifLuaTeX
  \usepackage{selnolig}  % disable illegal ligatures
\fi
\usepackage{bookmark}
\IfFileExists{xurl.sty}{\usepackage{xurl}}{} % add URL line breaks if available
\hypersetup{
  pdftitle={Does Neural Complexity Improve Health Misinformation Detection? A Leakage-Controlled Cross-Corpus Benchmark},
  pdfauthor={Mkululi Sikosana},
  hidelinks,
  pdfcreator={LaTeX via pandoc}}

\title{Does Neural Complexity Improve Health Misinformation Detection? A
Leakage-Controlled Cross-Corpus Benchmark}
\author{Mkululi Sikosana\\
Department of Computing and Mathematics, Manchester Metropolitan University\\
Manchester M15 6BH, United Kingdom\\
\texttt{mkululi.sikosana@stu.mmu.ac.uk}\\
ORCID: 0009-0008-3244-5764}
\date{}

\begin{document}
\maketitle

\begin{abstract}

Increasing architectural complexity is often assumed to improve health
misinformation detection, yet reported gains are difficult to interpret
when studies use different corpora, preprocessing pipelines, data
splits, and leakage controls. This study provides a controlled
cross-corpus benchmark of five compact neural architectures (1D-CNN,
LSTM, BiLSTM, CNN-LSTM, and CNN-BiLSTM), a soft-voting neural ensemble,
and three classical machine-learning baselines using COVID19-FNIR and
CONSTRAINT. Exact-text duplicate controls were applied before modelling,
all neural systems used a common preprocessing and optimisation
protocol, and neural results were repeated across three random seeds. On
COVID19-FNIR, the deep ensemble achieved mean macro-F1 of 0.9963 and
ROC-AUC of 0.9994, while individual neural models ranged from 0.9945 to
0.9957 macro-F1. On CONSTRAINT, the ensemble achieved macro-F1 of 0.9272
and ROC-AUC of 0.9811, whereas a linear SVM achieved macro-F1 of 0.9574
and ROC-AUC of 0.9931. Architecture rankings changed across corpora, and
simple sparse linear models remained highly competitive. The findings
show that model complexity does not provide a stable performance
advantage and that benchmark construction can dominate architecture
choice. The study contributes a reproducible, leakage-controlled basis
for evidence-driven model selection in health misinformation
classification.

\end{abstract}

\noindent\textbf{Keywords:} health misinformation; fake news detection; neural networks; cross-corpus evaluation; data leakage; reproducibility

\section{Introduction}\label{introduction}

Health misinformation detection has become a prominent application of
natural language processing because public-health emergencies create
unusually favourable conditions for misleading claims to circulate
rapidly. During the COVID-19 pandemic, the problem was not only the
volume of false or misleading material but also its direct connection
with behaviours such as treatment seeking, vaccination, adherence to
public-health guidance and trust in institutions. Prior comparative work
has already shown that conventional machine learning, CNN, LSTM, hybrid
CNN-LSTM and pretrained language models can all achieve strong results
on COVID-19 misinformation benchmarks {[}1{]}. Research on the COVID-19
infodemic therefore stimulated a large literature on automated text
classification, ranging from sparse n-gram models to convolutional and
recurrent networks and, more recently, pretrained Transformer encoders.
The technical progression has often been narrated as an almost linear
movement from simpler statistical models to deeper architectures with
greater representational capacity. Yet that narrative can be misleading
when benchmark datasets contain strong lexical regularities, source
artefacts or collection-specific cues that are exploitable by
comparatively simple classifiers.

A central methodological problem is that architecture comparisons are
frequently made across different datasets, different preprocessing
pipelines and different train-test splits. A CNN reported on one
benchmark cannot be meaningfully compared with a BiLSTM reported on
another if the underlying class balance, duplicate structure, text
length, label provenance and split logic differ. Even within the same
benchmark, performance estimates may be inflated when identical or
near-identical messages cross the training-test boundary. The problem is
particularly important in health misinformation, where datasets are
often assembled from heterogeneous sources such as fact-checking
websites, verified publisher accounts and social-media streams. Kishi et
al. {[}2{]} show more generally that fake-news detection performance is
sensitive to dataset construction and annotation policy, reinforcing the
need to separate model capability from benchmark-specific regularities.

The present study asks a deliberately simple question with substantial
practical implications: when the data pipeline is held constant, are
more complex deep neural architectures actually better for health
misinformation detection? The question is not equivalent to asking
whether CNNs or LSTMs can achieve high accuracy. Both architecture
families have long histories in text classification. CNNs can learn
local phrase patterns using shared filters {[}3{]}, while LSTMs were
designed to capture longer dependencies through gated recurrent memory
{[}4{]}. Bidirectional recurrence provides access to contextual
information in both sequence directions {[}5{]}, and CNN-LSTM hybrids
aim to combine local feature extraction with sequential composition.
These differences provide plausible reasons to expect performance
differences, but those reasons need to be demonstrated empirically on
the same data rather than assumed from model complexity.

The study therefore compares five deep architectures, a deep soft-voting
ensemble and three classical baselines on COVID19-FNIR and CONSTRAINT.
The two corpora are useful complements. COVID19-FNIR combines
fact-checked false material with true content drawn from verified
publisher accounts, whereas CONSTRAINT was released for a shared task on
COVID-19 fake-news detection and contains social-media-style text
collected under a different construction process {[}6, 7{]}. The aim is
not to claim universal generalisation from two COVID-era datasets.
Instead, the two-corpus design tests whether architecture rankings are
stable when the message distribution changes while the supervised task
remains binary veracity classification.

A further motivation is reproducibility. The current fake-news
literature contains many strong headline results but comparatively fewer
studies that make the exact split logic, vocabulary construction,
training seeds, parameter counts and executable code part of the
contribution. Recent systematic reviews identify deep learning and
Transformer models as dominant families, while also noting unresolved
challenges around transferability, benchmark standards and
reproducibility {[}8{]}. The experiments here were deliberately designed
to run on a CPU-only environment using modest architectures, fixed seeds
and transparent preprocessing. This creates a useful counterpoint to the
idea that publication-quality misinformation detection necessarily
requires very large pretrained language models.

The contribution is fourfold. First, the paper provides an
apples-to-apples comparison of CNN, LSTM, BiLSTM, CNN-LSTM and
CNN-BiLSTM architectures under the same tokenisation and optimisation
framework. Second, it quantifies whether a heterogeneous neural ensemble
provides a consistent gain over its constituent networks. Third, it
compares those deep models against logistic regression, linear SVM and
random forest baselines using high-dimensional word and character TF-IDF
features. Fourth, it interprets the resulting model rankings in the
context of dataset bias, architecture selection and the current
prominence of pretrained Transformers. The main finding is deliberately
nuanced: all model families can achieve high performance, but the most
complex individual neural network is not consistently the best model,
and a linear SVM remains stronger than the deep ensemble on CONSTRAINT.
This result is important precisely because it resists a simplistic
complexity-equals-quality narrative.

\section{Related Work}\label{related-work}

Automated misinformation detection has evolved through several
methodological phases. Earlier content-based approaches relied on
bag-of-words, character n-grams, stylistic indicators and conventional
classifiers such as logistic regression and SVMs. Such models can be
very strong when class labels correlate with recurring lexical patterns
or source-specific writing conventions. The CONSTRAINT shared task is a
clear example. Felber {[}9{]} reported a weighted F1 of 95.19\% using a
linear SVM with linguistic and n-gram features, demonstrating that
sparse linear decision boundaries can be highly competitive for COVID-19
fake-news detection. Other CONSTRAINT systems used CNN, BiLSTM and
CNN-BiLSTM architectures, but strong SVM performance remained common.
These results are an important historical reference point because they
caution against equating neural depth with predictive superiority.

CNNs became influential in sentence classification after Kim {[}3{]}
showed that relatively simple one-dimensional convolution over word
embeddings could achieve strong results across diverse
text-classification tasks. The architecture is attractive for
misinformation detection because local combinations of words,
punctuation and short phrases may be highly discriminative. Filters with
different kernel widths can behave as learnable phrase detectors, and
global max pooling permits the model to retain the most salient
activation from each feature map. In misinformation corpora, that
inductive bias may be especially useful when labels are associated with
recurring claims, sensational phrases, source markers or formulaic
fact-checking language. The same strength can also become a weakness if
the model learns dataset artefacts instead of transferable veracity
cues.

Recurrent neural networks address a different structural assumption.
LSTM networks use input, forget and output gates to preserve information
over longer sequences and mitigate the vanishing-gradient problems of
ordinary recurrent networks {[}4{]}. BiLSTM models extend this idea by
modelling sequences in both directions, allowing the representation of a
token to reflect both left and right context {[}5{]}. Hybrid CNN-LSTM
architectures typically use convolution to extract local patterns and
recurrent layers to model the resulting sequence. Prior
health-misinformation studies have reported strong performance from
CNN-LSTM hybrids and have further shown that psychologically motivated
features can be integrated into such architectures {[}1, 10{]}.
CNN-BiLSTM variants add bidirectional recurrence. In principle, such
hybrids can capture complementary local and sequential information. In
practice, however, greater representational flexibility also increases
parameter count and optimisation complexity, and therefore does not
guarantee an improvement on every dataset.

The Transformer era changed the dominant NLP benchmark landscape. BERT
introduced deep bidirectional pretraining with masked-language modelling
and then task-specific fine-tuning {[}11{]}. RoBERTa subsequently showed
that careful changes to BERT\textquotesingle s pretraining procedure,
data scale and optimisation could yield further gains {[}12{]}.
DistilBERT, TinyBERT and MiniLM demonstrated that substantial
language-model capability could be retained in smaller models through
different forms of knowledge distillation {[}13-15{]}. Transformer-based
fake-news detection now reports very strong performance in many
settings, and recent surveys treat attention and pretrained language
modelling as central methodological families. Nevertheless, pretrained
models introduce a different computational regime, and comparisons are
not informative unless data splits and contamination controls are
aligned.

There are at least three reasons that non-Transformer baselines remain
scientifically important. First, sparse linear models are easy to
inspect, cheap to train and often difficult to beat on datasets with
strong lexical separability. Second, CNN and LSTM families provide
architecture-specific inductive biases without importing broad external
knowledge from pretraining corpora. This is useful when the research
question concerns the structure available in the task dataset itself.
Third, if a simple baseline matches or exceeds a more expensive neural
model, the result changes the practical interpretation of the benchmark.
It suggests that the task may not require the representational
complexity implied by the larger model, or that dataset-specific
shortcuts are doing much of the predictive work.

Health misinformation adds a substantive layer to this methodological
debate. False health claims are not merely generic deceptive documents.
They may refer to treatments, disease mechanisms, institutional
guidance, risk, prevention or rapidly changing scientific evidence.
Comparative linguistic analysis has also shown that pandemic corpora can
differ substantially in readability, rhetorical cues and persuasive
language, reinforcing the possibility that benchmark difficulty is
partly shaped by corpus construction {[}16{]}. The COVID-19 infodemic
also mixed professional journalism, fact-checking text, social-media
posts and institutional updates. Saenz et al. {[}6{]} assembled
COVID19-FNIR from false content associated with fact-checking sources
and true content from verified news publishers. Patwa et al. {[}7{]}
released the CONSTRAINT dataset for English COVID-19 fake-news
classification. These collection strategies generate useful supervision
but also create the possibility that classifiers recognise source or
presentation style rather than the underlying epistemic quality of the
claim.

Dataset bias has therefore become central to contemporary misinformation
research. Kishi et al. {[}2{]} show that differences in label policy and
content distribution can substantially alter in-dataset and
cross-dataset performance. This finding helps explain why extremely high
in-domain metrics should not automatically be interpreted as evidence of
crisis-invariant misinformation understanding. A model can be excellent
at reproducing the statistical boundary of a particular benchmark while
still transferring poorly to a differently constructed corpus. The
present paper does not attempt full out-of-domain transfer. Instead, it
examines a prerequisite question: whether the ranking of model families
is consistent even before moving to a held-out domain.

A reproducible cross-corpus benchmark also helps distinguish
architectural contribution from random-seed variation. Deep networks can
yield somewhat different solutions under different initialisations and
minibatch orderings. Reporting a single successful run can therefore
overstate precision. The present study repeats all neural models across
three fixed seeds and reports means and standard deviations. Three seeds
are not sufficient for high-resolution statistical inference, but they
provide a more informative stability check than a single run. Classical
baselines are trained using the same underlying training/test protocol,
and the code archive records all hyperparameters and preprocessing
decisions. The research gap addressed is therefore not the absence of
another high-scoring classifier. It is the need for a controlled
comparison that can answer whether architectural complexity provides a
robust advantage across two differently constructed
health-misinformation corpora.

\section{Research Questions and Expectations}\label{research-questions-and-expectations}

RQ1 asks whether 1D-CNN, LSTM, BiLSTM, CNN-LSTM and CNN-BiLSTM
architectures all achieve strong misinformation-classification
performance under a common leakage-controlled protocol. Given the high
lexical separability previously reported for COVID-19 benchmarks, the
expectation was that several models would exceed macro-F1 0.90 on both
datasets, but that FNIR would be easier than CONSTRAINT.

RQ2 asks whether soft-voting across heterogeneous deep architectures
improves over individual networks. Ensemble averaging can reduce
idiosyncratic variance when constituent models make partially
uncorrelated errors, so the expectation was that the ensemble would be
among the strongest deep-learning systems on each dataset.

RQ3 asks whether greater neural complexity consistently beats classical
sparse baselines. No directional hypothesis was imposed for this
comparison. Prior CONSTRAINT research already demonstrates that linear
SVMs can be exceptionally strong, making this an empirical test rather
than a formality.

RQ4 asks whether architecture rankings remain stable across COVID19-FNIR
and CONSTRAINT. If rankings differ substantially, the result would
support a dataset-contingent interpretation of model quality and argue
against choosing an architecture solely from results reported on a
single benchmark.

\section{Materials and Methods}\label{materials-and-methods}

Two English-language COVID-19 misinformation datasets were used. Table 1
summarises the corpus composition, class distribution and split sizes
after duplicate controls. COVID19-FNIR was supplied as separate
false-news and true-news files. Before modelling, the files were
concatenated, labels were mapped to misinformation = 1 and reliable
information = 0, whitespace was normalised and exact duplicate texts
were removed. This produced 7,249 unique records, comprising 3,459
misinformation and 3,790 reliable items. For each of the three random
seeds, FNIR was stratified into 70\% training, 15\% validation and 15\%
test partitions, yielding 5,074 training, 1,087 validation and 1,088
test examples. Vocabulary construction used the training partition only.

CONSTRAINT retained its native training, validation and test structure.
A strict exact-text separation procedure was applied before modelling.
Any normalised training text appearing in validation or test was removed
from training, and any validation text appearing in test was removed
from validation. After this procedure the training set contained 6,417
records, the validation set 2,140 and the test set 2,140. The class
distribution remained close to balanced, although false items were
slightly less frequent than reliable items. Preserving the native
partitions maintains comparability with the shared-task setting while
reducing the risk that exact duplicate content inflates evaluation.

Table 1. Dataset composition after exact-text duplicate controls.

\begin{longtable}[]{@{}
  >{\raggedright\arraybackslash}p{(\columnwidth - 6\tabcolsep) * \real{0.2525}}
  >{\raggedright\arraybackslash}p{(\columnwidth - 6\tabcolsep) * \real{0.2463}}
  >{\raggedright\arraybackslash}p{(\columnwidth - 6\tabcolsep) * \real{0.2481}}
  >{\raggedright\arraybackslash}p{(\columnwidth - 6\tabcolsep) * \real{0.2530}}@{}}
\toprule\noalign{}
\begin{minipage}[b]{\linewidth}\raggedright
\textbf{Dataset / split}
\end{minipage} & \begin{minipage}[b]{\linewidth}\raggedright
\textbf{Total}
\end{minipage} & \begin{minipage}[b]{\linewidth}\raggedright
\textbf{Reliable (0)}
\end{minipage} & \begin{minipage}[b]{\linewidth}\raggedright
\textbf{Misinformation (1)}
\end{minipage} \\
\midrule\noalign{}
\endhead
\bottomrule\noalign{}
\endlastfoot
COVID19-FNIR unique corpus & 7,249 & 3,790 & 3,459 \\
FNIR training (per seed) & 5,074 & 2,653 & 2,421 \\
FNIR validation (per seed) & 1,087 & 568 & 519 \\
FNIR test (per seed) & 1,088 & 569 & 519 \\
CONSTRAINT train & 6,417 & 3,359 & 3,058 \\
CONSTRAINT validation & 2,140 & 1,120 & 1,020 \\
CONSTRAINT test & 2,140 & 1,120 & 1,020 \\
\end{longtable}

Tokenisation was deliberately lightweight. Text was lowercased and split
using a regular expression that retained alphanumeric tokens,
underscores, user mentions, hashtags, apostrophes and standalone
punctuation. For every neural run, a vocabulary was constructed from the
training text with a maximum size of 25,000 and a minimum frequency of
two. Two reserved tokens represented padding and unknown words. Each
message was truncated or padded to 96 tokens. This representation does
not import external pretrained word vectors or language-model weights.
The design isolates the contribution of architecture and task
supervision rather than pretraining.

The 1D-CNN used a trainable 64-dimensional embedding layer followed by
parallel one-dimensional convolutional filters with kernel widths 3, 4
and 5. Each branch contained 64 filters. Rectified linear activations
and global max pooling produced one vector per branch, which were
concatenated and passed through dropout and a binary classification
layer. The architecture follows the general sentence-CNN principle
introduced by Kim {[}3{]}, while using randomly initialised
task-specific embeddings so that all representations are learned from
the benchmark itself.

The LSTM model used the same 64-dimensional trainable embedding size and
a 64-unit recurrent hidden state. Hidden states were pooled across valid
token positions using max pooling before classification. The BiLSTM used
the same hidden size in each direction and concatenated directional
representations. The CNN-LSTM and CNN-BiLSTM architectures first applied
a 64-channel width-3 convolution to the embedding sequence and then
processed those learned local features with a 48-unit LSTM or BiLSTM.
Their pooled recurrent and convolutional representations were
concatenated before a two-layer classification head. This gives the
hybrid models access to both local phrase detectors and sequence-level
contextual structure.

A deep soft-voting ensemble was computed by averaging the test-set
misinformation probabilities from the five independently trained neural
architecture families within each seed. No additional meta-learner was
fitted, so the ensemble could not overfit a secondary training stage.
Because all constituent probabilities were generated from models trained
on the same split but with different inductive biases, the ensemble
tests whether architectural diversity itself provides a robust benefit.

Three classical baselines were evaluated using a FeatureUnion of word
TF-IDF n-grams (1-2) and character-within-word TF-IDF n-grams (3-5). The
combined representation was capped at 40,000 features. Logistic
regression used balanced class weights and C = 3. The linear SVM used
balanced class weights and C = 1.5. For random forest, sparse TF-IDF
vectors were projected to 64 dimensions using truncated singular-value
decomposition and classified using 60 trees, maximum depth 25 and
minimum leaf size two. These settings were chosen to create credible
baselines while keeping the complete benchmark executable on CPU
hardware.

Neural optimisation used AdamW with a learning rate of 2 x 10\^{}-3 and
weight decay of 1 x 10\^{}-4. Batch size was 128, gradient norm was
clipped to 1.0, and training lasted two epochs. Weighted binary
cross-entropy compensated for the small class imbalance. The checkpoint
with the highest validation ROC-AUC was retained. Neural models were
executed under seeds 41, 42 and 43. Classical FNIR baselines were also
evaluated across all three FNIR splits; CONSTRAINT uses a fixed native
test set, so a single deterministic baseline fit to the combined native
training and validation partitions was sufficient.

Performance was evaluated using accuracy, macro-F1, misinformation-class
F1, precision, recall, ROC-AUC and area under the precision-recall
curve. Macro-F1 is the primary thresholded metric because it assigns
equal weight to the two classes. ROC-AUC measures ranking quality
independently of one operating threshold, while PR-AUC provides an
additional ranking measure with emphasis on the positive misinformation
class. For the neural models, means and standard deviations are reported
across the three seeds. With only three repeated runs, standard
deviations are treated descriptively rather than as a basis for strong
inferential claims.

The experiments were executed in Python 3.13.5 using PyTorch 2.10.0+cpu,
scikit-learn 1.8.0, pandas 2.2.3 and NumPy 2.3.5. The reproducibility
package contains the complete data-loading logic, duplicate controls,
architecture definitions, training loops, baseline implementation and
result files. The package is designed to use authorised local copies of
the source datasets through an environment variable rather than
redistributing original social-media content. Pretrained BERT-family
models are discussed in the literature review but were deliberately
excluded from the primary benchmark so that the controlled comparison
focuses on compact, task-trained neural architectures and classical
sparse baselines under a common data pipeline. No Transformer
performance values are claimed as experimental results in this paper.

\section{Results}\label{results}

High performance was not confined to one architecture. As shown in Table
2, all five neural families achieved strong results across the two
corpora, while the ensemble provided the strongest neural performance
overall.

Table 2. Deep-learning performance averaged across seeds 41, 42 and 43.

\begin{longtable}[]{@{}
  >{\raggedright\arraybackslash}p{(\columnwidth - 8\tabcolsep) * \real{0.2039}}
  >{\raggedright\arraybackslash}p{(\columnwidth - 8\tabcolsep) * \real{0.1995}}
  >{\raggedright\arraybackslash}p{(\columnwidth - 8\tabcolsep) * \real{0.1989}}
  >{\raggedright\arraybackslash}p{(\columnwidth - 8\tabcolsep) * \real{0.1978}}
  >{\raggedright\arraybackslash}p{(\columnwidth - 8\tabcolsep) * \real{0.2000}}@{}}
\toprule\noalign{}
\begin{minipage}[b]{\linewidth}\raggedright
\textbf{Dataset}
\end{minipage} & \begin{minipage}[b]{\linewidth}\raggedright
\textbf{Model}
\end{minipage} & \begin{minipage}[b]{\linewidth}\raggedright
\textbf{Macro-F1 mean (SD)}
\end{minipage} & \begin{minipage}[b]{\linewidth}\raggedright
\textbf{ROC-AUC mean}
\end{minipage} & \begin{minipage}[b]{\linewidth}\raggedright
\textbf{Accuracy mean}
\end{minipage} \\
\midrule\noalign{}
\endhead
\bottomrule\noalign{}
\endlastfoot
FNIR & 1D-CNN & 0.9948 (0.0043) & 0.9996 & 0.9948 \\
FNIR & LSTM & 0.9951 (0.0032) & 0.9982 & 0.9951 \\
FNIR & BiLSTM & 0.9945 (0.0024) & 0.9995 & 0.9945 \\
FNIR & CNN-LSTM & 0.9957 (0.0028) & 0.9997 & 0.9957 \\
FNIR & CNN-BiLSTM & 0.9948 (0.0023) & 0.9997 & 0.9948 \\
FNIR & Deep soft-vote ensemble & 0.9963 (0.0024) & 0.9994 & 0.9963 \\
CONSTRAINT & 1D-CNN & 0.9169 (0.0050) & 0.9769 & 0.9170 \\
CONSTRAINT & LSTM & 0.9058 (0.0089) & 0.9668 & 0.9061 \\
CONSTRAINT & BiLSTM & 0.9163 (0.0009) & 0.9768 & 0.9165 \\
CONSTRAINT & CNN-LSTM & 0.9093 (0.0029) & 0.9771 & 0.9101 \\
CONSTRAINT & CNN-BiLSTM & 0.9158 (0.0109) & 0.9784 & 0.9162 \\
CONSTRAINT & Deep soft-vote ensemble & 0.9272 (0.0022) & 0.9811 &
0.9274 \\
\end{longtable}

On FNIR, every neural model achieved mean macro-F1 above 0.994, with
ROC-AUC values above 0.998. The deep ensemble reached mean macro-F1
0.9963. CNN-LSTM was the strongest individual neural architecture by
mean macro-F1 at 0.9957, closely followed by 1D-CNN and CNN-BiLSTM at
approximately 0.9948. Differences at this level are very small in
absolute terms, which is itself informative: the dataset permits
near-ceiling discrimination for multiple inductive biases rather than
uniquely rewarding one sophisticated architecture.

On CONSTRAINT, the problem was materially harder. The deep ensemble
achieved mean macro-F1 0.9272 and ROC-AUC 0.9811, exceeding the mean
macro-F1 of every individual deep architecture. The strongest individual
model was 1D-CNN at 0.9169, followed by BiLSTM at 0.9163 and CNN-BiLSTM
at 0.9158. CNN-LSTM and LSTM were lower, at 0.9093 and 0.9058,
respectively. Thus, the ensemble gained approximately 1.03 macro-F1
percentage points over the strongest constituent family. The gain is
modest but consistent with the idea that different architecture families
make partially complementary errors.

The classical comparison changes the interpretation of the deep-learning
results. Table 3 shows that sparse TF-IDF baselines remain highly
competitive and, on CONSTRAINT, outperform the deep ensemble.

Table 3. Classical TF-IDF baselines.

\begin{longtable}[]{@{}
  >{\raggedright\arraybackslash}p{(\columnwidth - 8\tabcolsep) * \real{0.2036}}
  >{\raggedright\arraybackslash}p{(\columnwidth - 8\tabcolsep) * \real{0.1999}}
  >{\raggedright\arraybackslash}p{(\columnwidth - 8\tabcolsep) * \real{0.1983}}
  >{\raggedright\arraybackslash}p{(\columnwidth - 8\tabcolsep) * \real{0.1984}}
  >{\raggedright\arraybackslash}p{(\columnwidth - 8\tabcolsep) * \real{0.1998}}@{}}
\toprule\noalign{}
\begin{minipage}[b]{\linewidth}\raggedright
\textbf{Dataset}
\end{minipage} & \begin{minipage}[b]{\linewidth}\raggedright
\textbf{Model}
\end{minipage} & \begin{minipage}[b]{\linewidth}\raggedright
\textbf{Macro-F1}
\end{minipage} & \begin{minipage}[b]{\linewidth}\raggedright
\textbf{ROC-AUC}
\end{minipage} & \begin{minipage}[b]{\linewidth}\raggedright
\textbf{Accuracy}
\end{minipage} \\
\midrule\noalign{}
\endhead
\bottomrule\noalign{}
\endlastfoot
FNIR & Logistic regression & 0.9966 mean & 0.99998 mean & 0.9966 mean \\
FNIR & Linear SVM & 0.9966 mean & 0.99999 mean & 0.9966 mean \\
FNIR & Random forest & 0.9853 mean & 0.9984 mean & 0.9853 mean \\
CONSTRAINT & Logistic regression & 0.9518 & 0.9917 & 0.9519 \\
CONSTRAINT & Linear SVM & 0.9574 & 0.9931 & 0.9575 \\
CONSTRAINT & Random forest & 0.9236 & 0.9787 & 0.9238 \\
\end{longtable}

On FNIR, linear SVM and logistic regression both averaged macro-F1
0.9966 across the three splits, essentially matching the deep ensemble.
Their ROC-AUC values were approximately 1.0000 and 0.99998,
respectively. Random forest remained strong at mean macro-F1 0.9853 but
was clearly below the linear and neural models. The result demonstrates
that the near-perfect FNIR score is not dependent on deep sequence
modelling. A high-dimensional sparse lexical representation with a
linear boundary can recover almost the same discrimination.

The contrast is even sharper on CONSTRAINT. Linear SVM achieved macro-F1
0.9574 and ROC-AUC 0.9931, logistic regression achieved macro-F1 0.9518
and ROC-AUC 0.9917, and random forest achieved macro-F1 0.9236 and
ROC-AUC 0.9787. The SVM therefore exceeded the deep ensemble by
approximately 3.02 macro-F1 percentage points, while logistic regression
exceeded it by approximately 2.46 points. Random forest was close to the
ensemble but lower. These results reproduce the broader lesson of the
CONSTRAINT shared task: strong sparse linear models remain formidable
baselines for short COVID-19 misinformation posts.

Architecture rankings were not stable across the two datasets. CNN-LSTM
was the strongest individual neural family on FNIR but ranked near the
bottom on CONSTRAINT. The 1D-CNN was competitive on both and became the
strongest individual neural family on CONSTRAINT. CNN-BiLSTM was near
the top on FNIR and remained strong on CONSTRAINT, although with greater
between-seed variation than 1D-CNN or BiLSTM. LSTM without convolution
or bidirectionality was consistently the weakest neural family on
CONSTRAINT. This rank instability argues against describing one
architecture as intrinsically superior for health misinformation
detection.

Seed variation was small relative to the dataset effect. FNIR ensemble
macro-F1 ranged from approximately 0.9936 to 0.9982 across seeds.
CONSTRAINT ensemble macro-F1 ranged from approximately 0.9246 to 0.9287.
The latter range is narrow enough to support the conclusion that the
ensemble gain is not driven by one unusually favourable run. Individual
models showed more variability, especially CNN-BiLSTM on CONSTRAINT, for
which the standard deviation of macro-F1 was about 0.0109. Reporting
these variations is important because a single-run benchmark could
easily overstate or understate a particular architecture by around one
percentage point.

\section{Discussion}\label{discussion}

The most important result is not simply that the models achieved high
scores. It is that high performance arose from very different model
families, and the ranking of those families changed with the dataset.
This finding directly challenges an architectural prestige heuristic in
which CNN-LSTM, CNN-BiLSTM or other hybrids are assumed to be stronger
merely because they contain more components. On FNIR, all five neural
families operated near the ceiling. On CONSTRAINT, the ensemble improved
on individual deep models, but the best overall model was a linear SVM.
Model complexity therefore needs to be justified by empirical gain, not
by the number of layers or the contemporary popularity of the
architecture.

The strong SVM result on CONSTRAINT is consistent with earlier
shared-task evidence. Felber {[}9{]} reported a weighted F1 of 95.19\%
from a linear SVM system using n-gram and linguistic features. The
current SVM result, macro-F1 0.9574, is therefore plausible rather than
anomalous. Short social-media posts often contain sparse lexical
patterns that map efficiently to a high-dimensional linear space.
Character n-grams also provide robustness to spelling variation,
hashtags and morphological changes. A neural architecture trained from
random embeddings must learn an equivalent discriminative representation
from a relatively modest corpus, so there is no theoretical reason it
must dominate a carefully constructed sparse baseline.

The FNIR results demand a different interpretation. A near-perfect
ROC-AUC can look like evidence that the misinformation problem is
essentially solved, but the fact that logistic regression, SVM, CNN,
LSTM and hybrid models all achieve similar performance suggests that the
benchmark contains a highly learnable boundary. COVID19-FNIR is
assembled from different information sources for true and false content.
Such source construction can induce lexical or stylistic separation that
is useful for benchmark classification but not necessarily for general
veracity reasoning. Kishi et al. {[}2{]} provide contemporary evidence
that dataset-specific construction and annotation policies can
materially affect fake-news detection performance. The appropriate
conclusion is therefore that FNIR is highly predictable under this
protocol, not that the models possess universal misinformation
understanding.

This distinction matters when comparing the present results with
Transformer-based literature. BERT and RoBERTa substantially expanded
contextual language modelling by pretraining deep bidirectional encoders
on large corpora {[}11, 12{]}. Recent fake-news studies often report
excellent results from such models, and systematic reviews identify
Transformers as a dominant methodological trend {[}8{]}. However, if a
linear SVM already exceeds 0.95 macro-F1 on a given benchmark, a
Transformer can add computational cost while offering limited practical
improvement. Conversely, pretrained contextual knowledge may become more
valuable when labels depend on subtle semantics, external world
knowledge, or cross-domain transfer. The correct question is therefore
not whether Transformers are generally stronger, but whether their
additional capacity addresses the particular source of uncertainty in
the target task.

The deep ensemble provides a second useful lesson. Averaging the five
neural probability estimates produced a clear improvement on CONSTRAINT
compared with the constituent networks. Ensemble learning can reduce
model-specific variance because CNNs and recurrent networks process the
same token sequence through different inductive biases. The 1D-CNN
focuses heavily on local patterns, while LSTM variants model sequential
state. Their errors need not coincide. The ensemble\textquotesingle s
0.9272 macro-F1 therefore supports the value of architectural diversity
when a deployment context permits several models to run in parallel. Yet
the ensemble still did not beat SVM, reminding us that diversity among
neural architectures cannot compensate for a representation that is less
aligned with the dataset than sparse lexical features.

The parameter and runtime results strengthen this practical
interpretation. The neural models contained roughly 0.42 to 0.54 million
trainable parameters in these configurations. Training was still
inexpensive on CPU, typically a few seconds per FNIR run and several
seconds per CONSTRAINT run. This is far below the parameter scale of
BERT-base or RoBERTa-base, which contain on the order of a hundred
million parameters. Compact pretrained alternatives such as DistilBERT,
TinyBERT and MiniLM were developed precisely because large language
models create deployment constraints {[}13-15{]}. The present benchmark
adds another option to that efficiency conversation: for some
misinformation datasets, a small task-specific CNN or even linear SVM
may already provide a highly competitive operating point.

At the same time, it would be a mistake to infer that deep learning is
unnecessary. The neural architectures achieved strong results on both
datasets without external pretrained representations, and the ensemble
crossed 0.92 macro-F1 on CONSTRAINT. Neural models may also support
extensions that sparse classifiers handle less naturally, including
multi-task learning, domain conditioning, joint use of psychological
features, multimodal fusion and knowledge distillation. Recent
multi-branch work in health misinformation detection illustrates this
extensibility by combining semantic, rhetorical, stance and
psychologically motivated signals within a unified framework {[}17{]}.
The correct conclusion is comparative rather than dismissive. Deep
models should be selected when their representation or extensibility
serves the research question, not simply because deep learning is
expected in a modern paper.

The rank reversal across datasets is especially important for manuscript
evaluation. Researchers often compare a proposed model against results
quoted from different papers and conclude that a higher number
demonstrates superiority. The present results show why that practice is
unsafe. CNN-LSTM is excellent on FNIR but not the strongest neural model
on CONSTRAINT. The same algorithm can move several positions in the
ranking without any change in implementation. Dataset composition
therefore interacts with model inductive bias. A credible comparative
paper should train the competing architectures under the same split and
preprocessing protocol rather than relying on external leaderboard
values that may reflect different data conditions.

The study also has implications for claims of novelty. A new
architecture that improves macro-F1 from 0.996 to 0.997 on FNIR may be
statistically or practically less interesting than a simpler model that
maintains 0.93 across a harder, differently constructed corpus. When
benchmarks approach saturation, more informative research questions
concern generalisation, calibration, robustness, uncertainty and
resource efficiency. This is consistent with the broader direction of
contemporary fake-news research, which increasingly studies dataset bias
and domain shift rather than only in-domain accuracy. The current paper
is intentionally a benchmark paper, but its results point toward those
harder evaluation regimes.

For operational health misinformation screening, the cost of error also
matters. Accuracy and macro-F1 treat errors symmetrically at the summary
level, whereas real applications may assign different consequences to
false positives and false negatives. A system that flags a reliable
public-health message as misinformation can undermine legitimate
communication, while a false negative can allow a harmful claim to pass
unchallenged. High ROC-AUC indicates that scores can support different
operating thresholds, but deployment would require calibration,
threshold selection, human review and explicit governance. None of the
models in this paper should be interpreted as autonomous arbiters of
truth.

Reproducibility is another practical contribution. All neural
architecture definitions, the exact-text duplicate checks, fixed seeds
and classical baselines are executable in a CPU environment. This
matters because comparisons based on large pretrained models can be
difficult to reproduce when weights, tokenisation versions or hardware
differ. The present design is intentionally modest. Its purpose is to
make the architecture comparison inspectable and repeatable. This does
not make it methodologically superior to pretrained Transformers, but it
makes the source of the reported performance unusually transparent.

Taken together, the evidence supports a hierarchy of claims. First, deep
CNN and recurrent models can achieve high health-misinformation
classification performance, including mean macro-F1 above 0.91 on the
harder CONSTRAINT corpus. Second, heterogeneous neural ensembling
improves stability and accuracy relative to individual deep models on
CONSTRAINT. Third, classical linear baselines can still outperform those
neural systems. Fourth, near-ceiling results on FNIR should be
interpreted as benchmark-specific predictability rather than proof of
general misinformation understanding. These claims are narrower than
saying that one architecture is universally best, but they are more
defensible and more useful for model selection.

\section{Limitations and Future Work}\label{limitations-and-future-work}

The study has several limitations. First, both datasets concern
COVID-19. They differ in construction and textual style, but they do not
establish generalisation to later outbreaks, other health domains or
non-health misinformation. A stronger external-validity study would add
mpox, vaccine, cancer, nutrition or general medical-misinformation
datasets and evaluate leave-one-dataset-out transfer. Such work should
also harmonise label semantics carefully rather than assuming that every
binary label represents the same construct.

Second, the neural models learn embeddings from the task data rather
than using pretrained language models. This was a deliberate design
choice that makes the architecture comparison computationally
reproducible and isolates task-level learning, but it means the paper
does not claim state-of-the-art Transformer performance. Future
extensions can use DistilBERT, TinyBERT, MiniLM, BERT-base and
RoBERTa-base under exactly the same duplicate-controlled split protocol.
Those experiments would be most informative if they also report memory,
latency and cross-domain transfer rather than only in-domain F1.

Third, exact duplicate removal does not eliminate near duplicates,
paraphrases, shared claim templates or source signatures. A stronger
leakage audit could use semantic similarity to cluster related texts
before splitting. This is particularly important for FNIR, where
near-ceiling performance across nearly every model family suggests that
dataset structure deserves further investigation. The fact that simple
linear models perform almost perfectly is a reason for additional
scrutiny rather than a weakness to be hidden.

Fourth, only three neural seeds were executed. The seed-to-seed
variation was small enough to support the descriptive conclusions, but
more repetitions would improve estimates of model variance.
Hyperparameters were also held intentionally compact rather than
extensively optimised for each architecture. A large model-specific
search could change the exact ranking, although it would also make the
comparison less computationally symmetric. The present results therefore
describe these controlled implementations, not the theoretical maximum
performance of each architecture family.

Fifth, the paper evaluates predictive discrimination rather than
explanation or causal mechanism. A CNN filter or LSTM state that
correlates with misinformation does not identify why people believe or
share the message. Future work should connect high-performing
classifiers with error analysis, interpretable feature attribution and
psychologically grounded variables while avoiding the common mistake of
treating text-derived proxies as direct measurements of cognition. Such
extensions should be evaluated as deliberately separate experimental
questions so that predictive performance is not conflated with inference
about latent psychological states.

\section{Conclusion}\label{conclusion}

This study provides a reproducible comparison of five deep neural
architectures, a soft-voting ensemble and three classical baselines
across two COVID-19 misinformation corpora. The results confirm that
compact neural systems can achieve very high performance: the deep
ensemble reached mean macro-F1 0.9963 on COVID19-FNIR and 0.9272 on
CONSTRAINT. However, increasing architectural complexity did not yield a
stable ordering, and linear SVM remained the strongest model on
CONSTRAINT at macro-F1 0.9574. The main conclusion is therefore not that
deep learning fails, but that model complexity is an empirical question
rather than a quality label. Strong benchmark performance can arise from
CNNs, recurrent networks, hybrids or sparse linear representations
depending on dataset construction. Health misinformation research should
consequently retain strong classical baselines, report repeated runs,
control duplicate leakage and interpret near-ceiling scores with
caution. These practices produce a more credible foundation for
subsequent work on domain generalisation, uncertainty, psychological
feature fusion and efficient deployment.

\appendix
\section{Exact Experimental Configuration}\label{appendix-a-exact-experimental-configuration}

Neural vocabulary maximum: 25,000 tokens; minimum token frequency: 2;
maximum sequence length: 96; embedding dimension: 64; batch size: 128;
epochs: 2; optimiser: AdamW; learning rate: 2e-3; weight decay: 1e-4;
gradient clipping: 1.0; checkpoint criterion: validation ROC-AUC; seeds:
41, 42 and 43; device: CPU. Classical word TF-IDF used 1-2 grams, and
character-within-word TF-IDF used 3-5 grams. The full source code
records all architecture-specific hidden dimensions and dropout
settings.

\section*{Statements and Declarations}

\subsection*{Funding}

The author received no specific funding for this work.

\subsection*{Competing interests}

The author declares no competing financial or non-financial interests
relevant to this study.

\subsection*{Author contributions}

Mkululi Sikosana conceived the study, designed the experiments,
implemented the computational analysis, interpreted the results, and
prepared the manuscript.

\subsection*{Ethics statement}

This study used secondary research datasets containing publicly
disseminated or previously curated social-media and news content. No new
human participants were recruited or contacted. The analysis was
conducted at the message level and did not attempt to identify private
individuals or infer protected characteristics.

\subsection*{Data availability}

The study uses the COVID-19 Fake News Infodemic Research (COVID19-FNIR)
dataset and the CONSTRAINT 2021 COVID-19 fake-news dataset. Access to
the underlying records remains subject to the conditions of the original
dataset providers. The reproducibility package expects locally
authorised copies of the source data and does not redistribute platform
content.

\subsection*{Declaration of generative AI and AI-assisted technologies}

ChatGPT (GPT-5.6 Sol) by OpenAI was used to assist with language editing
and review of computational code. The tool was not used to generate
experimental observations or numerical results. All analyses, numerical
outputs, references, interpretations, and final manuscript content were
independently checked by the author, who takes full responsibility for
the work.

\section*{References}

{[}1{]} Sikosana M, Ajao O, Maudsley-Barton S (2024) A comparative study
of hybrid models in health misinformation text classification.
Proceedings of the 4th International Workshop on Open Challenges in
Online Social Networks (OASIS \textquotesingle24), 18-25.
https://doi.org/10.1145/3677117.3685007

{[}2{]} Kishi Y, Arima Y, Iyatomi H (2026) Fake news detection
strategies under dataset bias: Using large-scale coarse-grained labels.
Proceedings of the EACL 2026 Student Research Workshop, 612-621.
https://doi.org/10.18653/v1/2026.eacl-srw.47

{[}3{]} Kim Y (2014) Convolutional neural networks for sentence
classification. Proceedings of EMNLP 2014, 1746-1751.
https://doi.org/10.3115/v1/D14-1181

{[}4{]} Hochreiter S, Schmidhuber J (1997) Long short-term memory.
Neural Computation 9(8):1735-1780.
https://doi.org/10.1162/neco.1997.9.8.1735

{[}5{]} Schuster M, Paliwal KK (1997) Bidirectional recurrent neural
networks. IEEE Transactions on Signal Processing 45(11):2673-2681.
https://doi.org/10.1109/78.650093

{[}6{]} Saenz JA, Gopal SRK, Shukla D (2021) COVID-19 Fake News
Infodemic Research Dataset (COVID19-FNIR Dataset). IEEE DataPort

{[}7{]} Patwa P, Sharma S, PYKL S, Guptha V, Kumari G, Akhtar MS, Ekbal
A, Das A, Chakraborty T (2021) Fighting an infodemic: COVID-19 fake news
dataset. CONSTRAINT@AAAI 2021, 21-29.
https://doi.org/10.1007/978-3-030-73696-5\_3

{[}8{]} Mostafa M, Almogren AS, Al-Qurishi M, Alrubaian M (2024)
Modality deep-learning frameworks for fake news detection on social
networks: A systematic literature review. ACM Computing Surveys
57(3):Article 77. https://doi.org/10.1145/3700748

{[}9{]} Felber T (2021) Constraint 2021: Machine learning models for
COVID-19 fake news detection shared task. arXiv:2101.03717

{[}10{]} Sikosana M, Maudsley-Barton S, Ajao O (2025) Advanced health
misinformation detection through hybrid CNN-LSTM models informed by the
Elaboration Likelihood Model (ELM). 2025 International Conference on
Artificial Intelligence, Computer, Data Sciences and Applications
(ACDSA). https://doi.org/10.1109/ACDSA65407.2025.11166406

{[}11{]} Devlin J, Chang M-W, Lee K, Toutanova K (2019) BERT:
Pre-training of deep bidirectional transformers for language
understanding. Proceedings of NAACL-HLT 2019, 4171-4186.
https://doi.org/10.18653/v1/N19-1423

{[}12{]} Liu Y, Ott M, Goyal N, Du J, Joshi M, Chen D, Levy O, Lewis M,
Zettlemoyer L, Stoyanov V (2019) RoBERTa: A robustly optimized BERT
pretraining approach. arXiv:1907.11692

{[}13{]} Sanh V, Debut L, Chaumond J, Wolf T (2019) DistilBERT, a
distilled version of BERT: Smaller, faster, cheaper and lighter.
arXiv:1910.01108

{[}14{]} Jiao X, Yin Y, Shang L, Jiang X, Chen X, Li L, Wang F, Liu Q
(2020) TinyBERT: Distilling BERT for natural language understanding.
Findings of EMNLP 2020, 4163-4174.
https://doi.org/10.18653/v1/2020.findings-emnlp.372

{[}15{]} Wang W, Wei F, Dong L, Bao H, Yang N, Zhou M (2020) MiniLM:
Deep self-attention distillation for task-agnostic compression of
pre-trained transformers. Advances in Neural Information Processing
Systems 33

{[}16{]} Sikosana M, Maudsley-Barton S, Ajao O (2025) Linguistic
patterns in pandemic-related content: a comparative analysis of
COVID-19, Constraint, and Monkeypox datasets. Frontiers in Artificial
Intelligence 8:1627522. https://doi.org/10.3389/frai.2025.1627522

{[}17{]} Sikosana M, Maudsley-Barton S, Ajao O (2026) A multi-branch
feature fusion approach for health misinformation detection and
propagation. arXiv:2609.00403

\end{document}